\documentclass[sigconf]{acmart}

\usepackage{subcaption}
\usepackage{algorithm2e}
\usepackage[capitalise]{cleveref}
\usepackage{wrapfig}

\usepackage{color}
\usepackage{xcolor}
\definecolor{darkred}{rgb}{0.6,0,0}
\definecolor{blue}{rgb}{0,0,0.75}

\AtBeginDocument{%
  \providecommand\BibTeX{{%
    \normalfont B\kern-0.5em{\scshape i\kern-0.25em b}\kern-0.8em\TeX}}}

\setcopyright{rightsretained}
\copyrightyear{2024}
\acmYear{2024}
\acmConference{SIGGRAPH Posters '24}{July 27 - August 01, 2024}{Denver, CO, USA}\acmBooktitle{Special Interest Group on Computer Graphics and Interactive Techniques Conference Posters (SIGGRAPH Posters '24), July 27 - August 01, 2024}\acmDOI{10.1145/3641234.3671025}
\acmISBN{979-8-4007-0516-8/24/07}

\begin{document}

\title{Image Segmentation from Shadow-Hints using Minimum Spanning Trees}

\author{Moritz Heep}
\email{mheep@uni-bonn.de}
\orcid{0009-0009-3760-8371}
\affiliation{%
  \institution{University of Bonn}
  \city{Bonn}
  \country{Germany}
}
\author{Eduard Zell}
\email{e.zell@hotmail.de}
\orcid{0009-0007-3467-9890}
\affiliation{%
  \institution{University of Bonn}
  \city{Bonn}
  \country{Germany}
}

\begin{teaserfigure}
    \centering
    \includegraphics[width=\linewidth, trim={3 359 33 3}, clip]{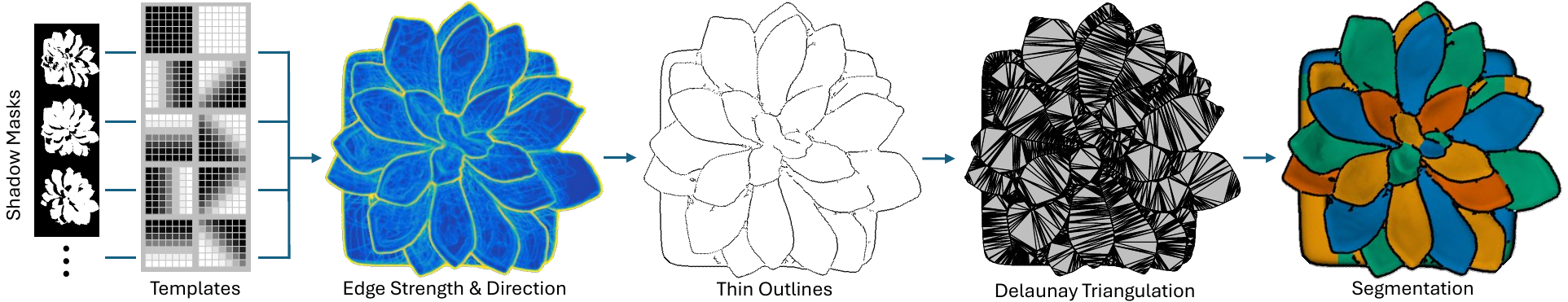}
    \caption{Overview of our Pipeline: Starting from a set of shadow masks, we use templates to extract light-to-shadow transitions. After combining these transitions into an edge strength and direction, we apply non-maximum suppression to obtain thin outlines. The segmentation is performed on a Delaunay triangulation of the detected outline points.}
    \Description{Overview of the Algorithm.}
  \label{fig:teaser}
\end{teaserfigure}

\maketitle
\section{Introduction}
Image segmentation in RGB space is a notoriously difficult task where state-of-the-art methods are trained on thousands or even millions of annotated images \cite{Kirillov2023}. While the performance is impressive, it is still not perfect. We propose a novel image segmentation method, achieving similar segmentation quality but without training. %
Instead, we require an image sequence with a static camera and a single light source at varying positions, as used in for photometric stereo, for example. 
Here, foreground objects cast shadows onto background objects, the detection of transitions from light to shadow can be used to reveal the spatial structure of the scene and to trace the contour of an object.
Unfortunately, these contours are not water-tight and simple flood-fill approaches fail. Inspired by interactive sketch colouring methods \cite{Parakkat2022}, our novel image segmentation algorithm is based on Delaunay triangulations. After converting the pixel grid to a mesh, our algorithm operates on the face graph of the Delaunay triangulation where there is no notion of colour similarity. Instead, we rely on the edge length as a similarity indicator due to the circumcircle property of the Delaunay triangulation. While comparable graph-based image segmentation algorithms \cite{Felzenszwalb2004} cluster pixels according to colour similarity and achieve - by current standards - only mediocre results, our method shows promising results without any training on annotated data.
\section{Method} %
Instead of RGB images, we require binary shadow masks for each light position to generate a discontinuity-sensitive image segmentation. Such shadow masks are commonly created in photometric stereo, see e.g.\ \cite{Heep2022}. 
\subsection{Shadow Edge Detection}
\label{sec:shadow_edge_detection}
We apply a template matching procedure to detect shadow-to-light transitions in all shadow masks and merge them into a pixel-based edge strength and direction. We use \emph{ten} templates (Fig.~\ref{fig:teaser}), each $7\times 7$ pixels big: Two for
completely lit or shadowed regions and eight for light-to-shadow transitions in the directions $d\in\mathcal{N}$ of the eight pixel neighbourhood. Let $E_{lp,0}$, $E_{lp,1}$ and $E_{lp,d}$ 
be the resulting $L^2$ errors at pixel $p$ under lighting $l$. %
Fully shadowed regions and transitions that cannot be explained by the current light position, cf.\ \cite{Raskar2004}, are excluded by setting the binary weight $\omega_{lp,d}\in\{0, 1\}$ to zero.
We calculate the edge score
\begin{align}
    b_{p,d}=\frac{\sum_l \omega_{lp,d}\cdot\sigma\big((E_{lp,1}-E_{lp,d})/\beta\big)}{\sum_l \omega_{lp,d}}\in[0,1]
\end{align}
for each direction $d\in\mathcal{N}$ and pixel $p$. The sigmoid function $\sigma$ smoothly transitions between $0$ and $1$, depending on whether a pixel is fully lit or the respective transition direction is a better fit. 
For each pixel, we choose the edge direction $\theta_p\in\mathcal{N}$ such that $b_{p,d}$ is maximal and use the maximum value $g_p$ as edge strength.
\subsection{Subpixel Delaunay Triangulation}
\label{sec:subpixel_delaunay}
To extract thin outlines from edge strength $g_p$ and direction $\theta_p$, we apply non-maximum suppression and double thresholding \cite{Canny1986}. Furthermore, we refine the location of each edge pixel, by employing a quadratic fit to locate the subpixel positions of the maxima in $g_p$. 
This yields smoother outlines and moves maxima away from pixel centres, i.e.\ pixels are unambiguously \emph{within} one segment. The maxima are then connected through a Delaunay triangulation, completing our shift from the pixel domain to a 2D polygon mesh. We remove any triangles covering the image background.
\subsection{Segmentation}
\label{sec:segmentation}
Our segmentation algorithm progressively fuses triangles into larger segments, cf.\ \cite{Kruskal1956}. Since vertices are more densely placed along detected outlines, the shared edge length between two triangles is a suitable proxy for how likely these two triangles belong to the same segment.
We exploit this by placing the edges in a priority queue with non-increasing edge length. Running any minimum spanning tree algorithm with these lengths would fuse all triangles into a single segment. Instead, we calculate the aspect ratio for each segment $S$ as the ratio between its area $|S|$ and the length $|e|$ of its shortest edge:
\begin{align}
    l_S:=\frac{|S|-A_\text{min}}{\min_{e\in S}(|e|)}\,.
\end{align}
Reducing the segment area by $A_{\text{min}}$ ensures that segments smaller than $A_{\text{min}}$ are fused in the further course of the algorithm. 
When processing edge $e$, we check if
\begin{align}
    |e|>\kappa\cdot\min\left(l_S,l_{S'}\right)
\end{align}
where $\kappa>0$ is a parameter to control the segment shape. 
If true, segments $S$ and $S'$ are fused along the edge $e$. 
Over time, the segments become larger and contain increasingly shorter edges. Hence, $l_S$ increases as segments grow until no more fusions occur. The algorithm terminates when all edges have been processed. 
Segments smaller than $A_{\text{min}}$ are always fused.
\section{Experiments}
\begin{figure}[tb]
        \includegraphics[width=\linewidth]{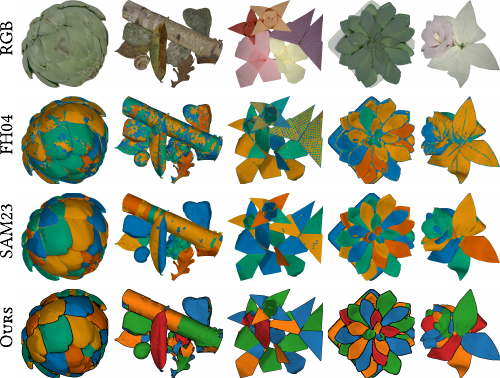}
        \caption{From Top to Bottom: RGB input used to generate segmentations with FH04 %
        and SAM23 %
        as well as our segmentation from shadow-hints. For our method, detected outline points are overlaid to visualize where completions occur.}
        \Description{Segmentation results for five different objects.}
        \label{fig:results}
\end{figure}
We tested our algorithm on different objects against classic and learning-based image segmentations: FH04 \cite{Felzenszwalb2004} 
operates on a nearest-neighbour graph built from screen position and RGB colour.
In contrast, the Segment Anything Model (SAM23) \cite{Kirillov2023} is a state-of-the-art deep learning approach.
FH04 leads to over-segmentation in textured regions (e.g.\ the wooden branch, Fig.~\ref{fig:results}) and under-segmentation for similarly coloured segments. 
SAM23 and our method are more robust and 
create comparable results in many cases. 
SAM23 is prone to over-segmentation in high-contrast textures (e.g.\ the origami butterfly, Fig.~\ref{fig:results}) while our approach is prone to under-segmentation if transitions between objects are too smooth to cast a shadow. 
\section{Conclusion}
Given the quality of the results without depending on annotated data, our method offers an alternative to create annotated datasets to train learning-based image segmentation algorithms. The granularity of the segmentation can be controlled in real-time through the user parameter $\kappa$ and manual refinement is feasible at the segment level, instead of the pixel level. 
\begin{acks}
    Special thanks to Amal Dev Parakkat for fruitful discussions.
    This work has been funded by the Deutsche Forschungsgemeinschaft (DFG, German Research Foundation) under Germany's Excellence Strategy, EXC-2070 -- 390732324 (PhenoRob).
\end{acks}
\bibliographystyle{ACM-Reference-Format}
\bibliography{siggraph24}
\end{document}